\documentclass[letterpaper, 10 pt, conference]{ieeeconf}  

\IEEEoverridecommandlockouts                              

\usepackage[utf8]{inputenc} 
\usepackage[T1]{fontenc}    
\usepackage{hyperref}       
\usepackage{url}            
\usepackage{booktabs}       
\usepackage{amsfonts}       
\usepackage{nicefrac}       
\usepackage{microtype}      
\usepackage{cleveref}  

\usepackage{graphicx} 
\usepackage{floatrow}
\usepackage{subfig}  
   \graphicspath{{figs/}}
   \DeclareGraphicsExtensions{.pdf,.jpg,.png,.eps}

\usepackage{siunitx}
\usepackage{multirow}

\usepackage{amsmath,amssymb}
\usepackage{xspace}

\newcolumntype{L}{S[table-format=2.1]@{\hskip 10pt}}
\newcolumntype{R}{S[table-format=2.1]}
\newcolumntype{?}{!{\vrule}}
\newcommand{\mytimes}{\medmuskip=0mu\times}
\newcommand{\dense}{\thickmuskip=2mu}

\renewcommand{\eqref}[1]{(\ref{#1})}
\newcommand{\figref}[1]{Fig.~\ref{#1}}

\newcommand{\ie}{\textrm{i.e.}\xspace}
\newcommand{\eg}{\textrm{e.g.}\xspace}
\newcommand{\etc}{\textrm{etc.}\xspace}
\newcommand{\etal}{\textrm{et~al.}\xspace}

\newcommand{\sss}{\scriptscriptstyle}

\newcommand{\model}{\ensuremath{M}\xspace}

\newcommand{\map}{\ensuremath{\mathbb{V}}\xspace}

\newcommand{\threed}{\mbox{3-D}\xspace}
\newcommand{\twod}{\mbox{2-D}\xspace}

\title{\LARGE \bf
Integrating Algorithmic Planning and Deep Learning \\ 
for Partially Observable Navigation
}

\author{Peter Karkus$^{1,2}$, David Hsu$^{1,2}$ and Wee Sun Lee$^{2}$
\thanks{$^{1}$NUS Graduate School for Integrative Sciences and Engineering}%
\thanks{$^{2}$School of Computing}%
\thanks{National University of Singapore, 119077, Singapore.}
\thanks{\tt{\{karkus, dyhsu, leews\}@comp.nus.edu.sg}}
} 

\begin{document}

\maketitle
\thispagestyle{empty}
\pagestyle{empty}


\begin{abstract}

We propose to take a novel approach to robot system design where each building block 
of a larger system is represented as a differentiable program, \ie{} a deep neural network. 
This representation allows for integrating algorithmic planning 
and deep learning in a principled manner, and thus combine the benefits of 
model-free and model-based methods. We apply the proposed approach to 
a challenging partially observable robot navigation task. The robot must navigate to a goal
in a previously unseen \threed environment without knowing its initial location, 
and instead relying on a \twod floor map and visual observations from an onboard camera.
We introduce the Navigation Networks (NavNets) that encode 
state estimation, planning and acting in a single, end-to-end trainable recurrent neural network.
In preliminary simulation experiments we successfully trained navigation networks to
solve the challenging partially observable navigation task.

\end{abstract}

\section{Introduction}

Humans employ two primal approaches to decision-making: reasoning with models
 and learning from experiences. The former is called planning, and the latter, learning. 
By integrating planning and learning methods, AlphaGo has recently achieved 
super-human performance in the challenging game of Go~\cite{silver2016mastering}.
Robots must also integrate planning and learning to address
truly difficult decision-making tasks, and ultimately to achieve human-level robotic intelligence. 

Robots act in the real world that is inherently uncertain and only partially observable. 
Under partial observability the robot cannot determine its
state exactly. Instead, it must integrate information over
the past history of its actions and observations. Unfortunately, this
drastically increases the complexity of decision making~\cite{papadimitriou1987complexity}.
 In the model-based approach, we may formulate the problem as a partially
  observable Markov decision process~(POMDP).
Approximate algorithms have made dramatic progress on solving
POMDPs~\cite{pineau2003applying,spaan2005perseus,kurniawati2008sarsop,silver2010monte,ye2017despot};
however, manually constructing POMDP models or learning them from data
remains difficult~\cite{littman2002predictive, shani2005model,
  boots2011closing}. In the model-free approach, we
 circumvent the difficulty of model construction by
directly searching for a solution within a policy class~\cite{baxter2001infinite,bagnell2003policy}. 
The key issue is then choosing priors
 that allow efficient policy search. 

Deep neural networks (DNNs) have brought unprecedented success in many
domains~\cite{krizhevsky2012imagenet, mnih2015human, silver2016mastering}. 
Priors on the network architecture make learning efficient, \eg{} convolutions
allow for local, spatially invariant features~\cite{lecun1989backpropagation}.  
DNNs provide a distinct new approach to partially observable
decision-making~\cite{hochreiter1997long, bakker2003robot, hausknecht2015deep, mirowski2016learning}.  
In particular, DQN, a convolutional architecture, has successfully tackled Atari games with
complex visual input~\cite{mnih2015human}.
Combining DQN with recurrent LSTM layers
allows them to deal with partial observability~\cite{hausknecht2015deep, mirowski2016learning}. 
However, such architectures are fundamentally limited, because
their priors fail to exploit the underlying sequential nature of planning.

\begin{figure}[!t]
  \centering
	\includegraphics[width=0.9\textwidth]{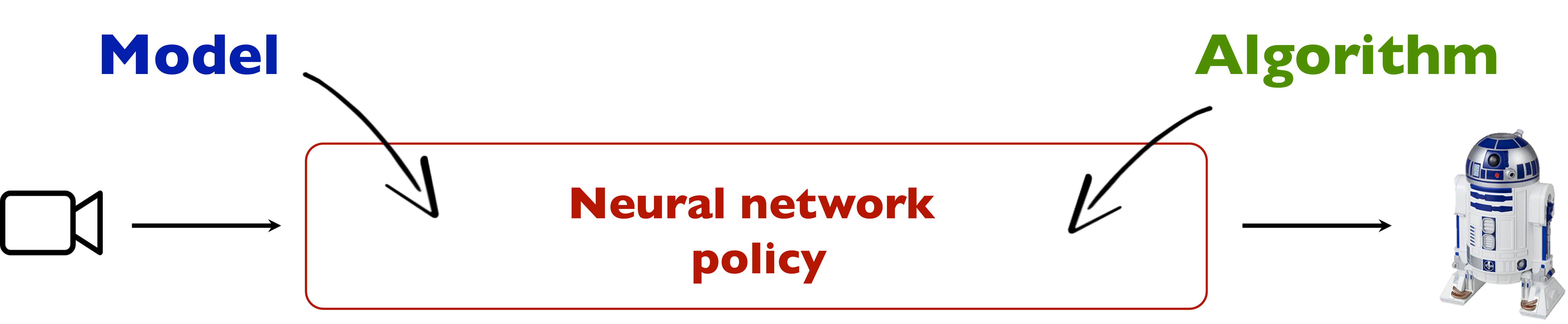}
  \caption{Integrating planning and learning through 
  algorithmic priors. The policy is represented by a model 
  connected to an algorithm that solves the model. 
 Both the model and the algorithm are 
  encoded in a single neural network. 
   \vspace{-12pt}}
\label{fig:model_algorithm}
\end{figure}

We want to combine the strength of algorithmic planning  and deep learning in 
order to scale to the challenges of real-world decision-making.
The question is then: how do we integrate the structure
of algorithmic planning into a deep learning framework, \ie{} 
what are the suitable priors on a DNN for decision making 
under partial observability?  
We propose \emph{algorithmic priors} that integrate algorithmic planning and 
deep learning by embedding both a model and an algorithm that solves the model, 
in a single DNN architecture (\figref{fig:model_algorithm}).

We apply this approach to a challenging partially observable navigation task, where a robot is placed in a previously
 unseen \threed maze and must navigate to a specified goal while avoiding unforeseen obstacles. 
 In this decision-making task the state of the robot is partially observable.
 The robot receives a \twod floor map, but
 it does not know its initial location on the map. Instead, it must infer it from long sequences 
 of sensor observations from an onboard monocular camera.  

We introduce the Navigation Networks (NavNet), a recurrent neural network (RNN) that employs 
algorithmic priors for navigation under partial observability. NavNets extend
our work on QMDP-nets that addressed
partially observable planning in simplified domains~\cite{karkus2017qmdp}.
In contrast to QMDP-nets, NavNets embed the entire solution structure of partially observable 
navigation: localization, planning and acting. Localization must now deal with camera images; 
and a reactive actor policy is now responsible for avoiding unforeseen obstacles.
In preliminary experiments NavNets successful learned policies from expert demonstrations
that generalize over simulated \threed environments. 
Results indicate that integrating algorithmic planning and deep learning enables 
reasoning with partial observations in large-scale, visual domains. In our experiments the robot 
is restricted to a grid, and actions are simple, discrete motion commands; however, the
 approach could be applied to continuous control in the future.


\section{Related Work}

The idea of embedding specific computation structure in a neural network
architecture has been gaining attention recently. 
Previous work addressed decision making in fully observable domains 
such as value iteration in Markov decision processes~\cite{tamar2016value},
searching in graphs~\cite{oh2017value, farquhar2017treeqn, guez2018learning}, 
optimal control~\cite{OkaRig17} 
and quadratic program optimization~\cite{amos2017optnet, donti2017task}.
These works do not address the issue of partial
observability which drastically increases the computational complexity of
decision making~\cite{papadimitriou1987complexity}.
Another group of work addressed probabilistic state 
estimation~\cite{haarnoja2016backprop, jonschkowski2016, karkus2018particle, jonschkowski2018differentiable},
but they do not deal with decision making or planning.  

Both Shankar~\etal{}~\cite{shankar2016reinforcement} and
Gupta~\etal{}~\cite{gupta2017cognitive} addressed planning under partial observability.
The former focuses on learning a model rather than a policy, where the model
does not generalize over environments.  The
latter proposes a network learning approach to robot navigation with the focus on mapping
instead of partially observable planning.
They address a navigation problem in unknown
environments where the location of the robot is always known.
In our setting the robot location is partially observable: it must be inferred from camera images 
and a \twod map.

Finally, we introduced QMDP-nets~\cite{karkus2017qmdp} that 
use algorithmic priors for learning partially observable decision making policies 
in simplified domains. 
In this work we extend QMDP-nets 
to tackle a much more complex navigation domain that requires 
reasoning with visual observations and accounting for unforeseen obstacles.


\section{Algorithmic priors}
We propose to use planning algorithms as ``priors'' on a DNN architecture 
for efficiently learning under partial observability.
We search for a policy in the form of a DNN, but impose the  
structure of both a model and a planning algorithm on the DNN 
architecture~(\figref{fig:model_algorithm}).
The weights of the neural network then correspond to the model parameters,
which are learned.
We do not want to rely on data alone for learning the 
 computational steps of planning. Instead, we 
 explicitly encode a planning algorithm in a network learning architecture, and thus
combine the strengths of model-free policy search and model-based planning.
The core idea for encoding an algorithm in a DNN is
 is to view neural networks as \emph{differentiable programs}, where 
 algorithmic operations are realized as layers of the neural network. 
For example, a weighted sum becomes a convolutional layer, a
maximum operation becomes a max-pool layer.

 We expect that embedding a model in the network allows efficient generalization
  over a large task space. Embedding an algorithm allows learning end-to-end and 
 thus circumvents the difficulties of traditional model-based learning. Moreover, 
  we may learn abstractions that compensate for the limitations of an approximate
  algorithm through end-to-end training~\cite{karkus2017qmdp}.   

Ultimately, we envision a fundamentally new approach to robotic systems 
design, where all building blocks 
are implemented as differentiable programs, and thus can be jointly optimized 
for the overall system objective.

\section{Navigation networks}

We define a partially observable navigation task that is prevalent
in mobile robot applications. A robot is placed in a previously unseen indoor 
environment and must navigate to a specified goal. The robot 
receives a \twod floor map that indicates walls and the 
goal; however, the robot does not know its own location on 
the map initially. Instead, it must estimate it based on past actions and 
observations from an onboard monocular camera. 
Although being uncertain of its location, the robot must choose actions that 
lead to the goal, while avoiding walls and other, previously unforeseen
 obstacles that are not indicated on its floor map, \eg, furniture.

 \begin{figure}[!t]
  \centering
	\includegraphics[width=0.9\textwidth]{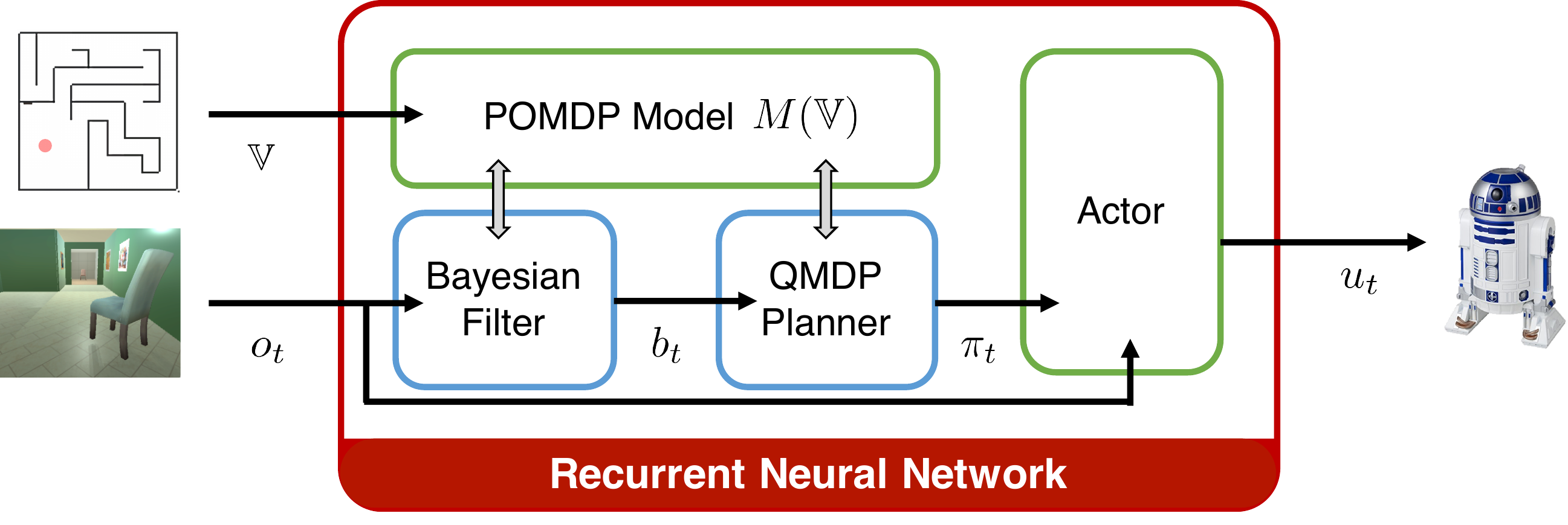}
  \caption{NavNets are recurrent neural networks with algorithmic priors. A NavNet 
  represents a policy, it maps from observations to actions, but it encodes  
 the entire solution structure of navigation -- state estimation, planning and acting --
 in a single, differentiable neural network.
   \vspace{-12pt}}
\label{fig:navnet}
\end{figure}

We introduce the Navigation Network, a deep RNN architecture that employs 
algorithmic priors specific to navigation under state uncertainty~(\figref{fig:navnet}).
Robot navigation is typically addressed by decomposing the problem to 
localization, planning and control. We apply the same decomposition, but 
encode all three components in a unified DNN representation.  
More specifically, the weights of a NavNet encode an abstract POMDP model, 
which is learned. The network also encodes an algorithm that solves the POMDP model.
First, a Bayesian histogram filter integrates information from sequences of 
visual observations and past actions
into a belief, \ie{} a probabilistic estimate of the state. Second, a 
QMDP planner creates a high-level plan given the map and the 
current belief.
Finally, an actor policy takes the high-level plan and a camera observation 
 and outputs an action. The actor 
generally chooses actions that execute the plan; however, it may 
also need to deviate from the plan to account for unforeseen obstacles 
blocking the path.

 \begin{figure*}[!t]
  \centering
	\includegraphics[height=2.6cm]{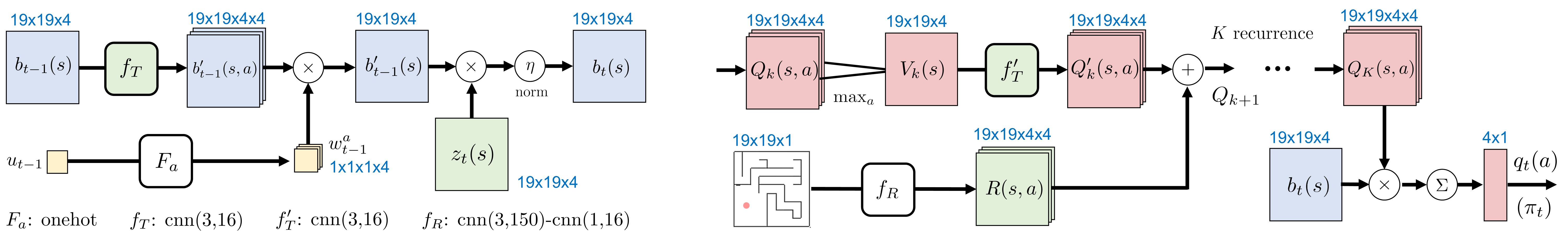}\vspace{-0pt}  
  \caption{Bayesian filter~(left) and QMDP planner~(right) encoded in the navigation network. 
  Belief tensors are in blue;
  learned POMDP components are in green; planned value tensors are in red; other tensors are in yellow.
  We use $\textrm{cnn}(k,f)$ to denote convolution with $k\mytimes k$ kernel and $f$ filters. 
  Activations and reshape operations are omitted for clarity.  
  The planner outputs action values, $q(a)$, which are then combined with a few steps of history and 
  fed to the actor policy show in~\figref{fig:controller}.
   \vspace{-6pt}} 
\label{fig:filterplanner}
\end{figure*}

 \begin{figure}[!t]
  \centering
	\includegraphics[height=2.8cm]{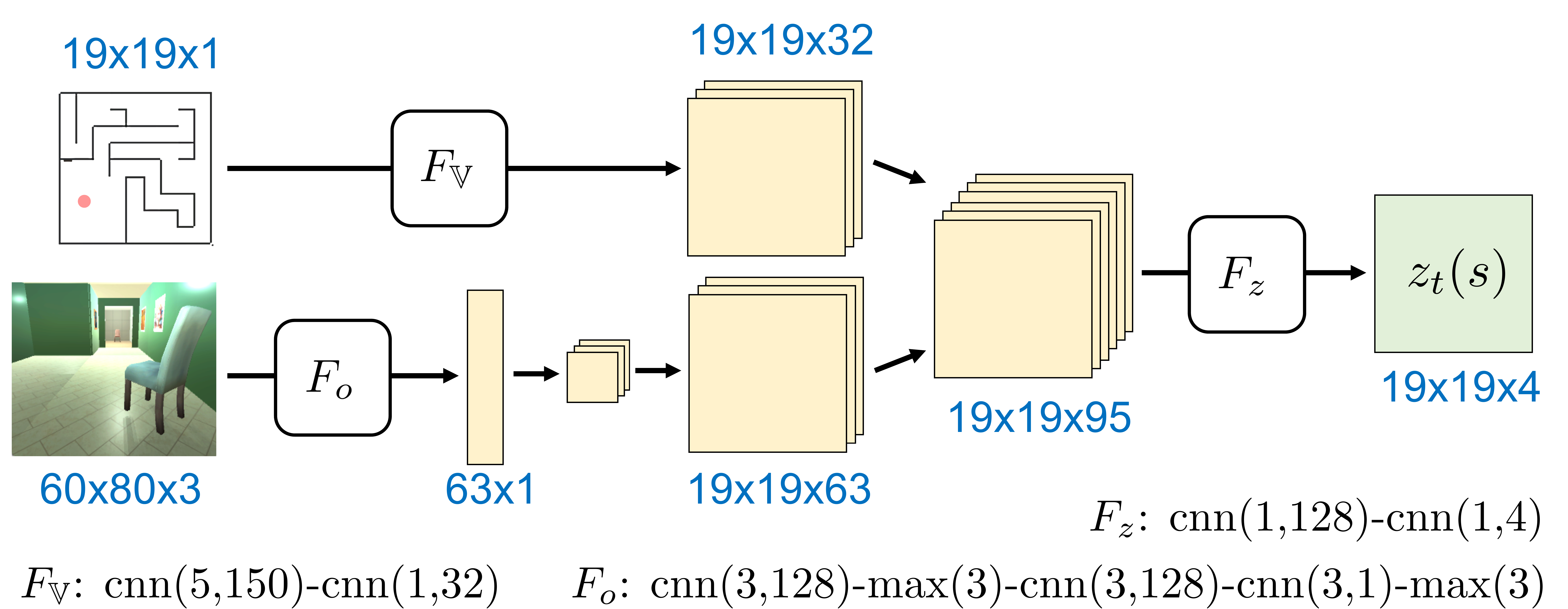}\vspace{-0pt}  
  \caption{Observation model that maps from a camera image $o_t$ and the floor map $\map$ to
  the distribution $z(s)=p(s | o_t, \map)$. We use  $\textrm{max}(k)$ to denote max-pooling with $k\mytimes k$ kernel. 
   \vspace{-12pt}} 
\label{fig:obsmodel}
\end{figure}

 \begin{figure}[!t]
  \centering
	\includegraphics[height=2.8cm]{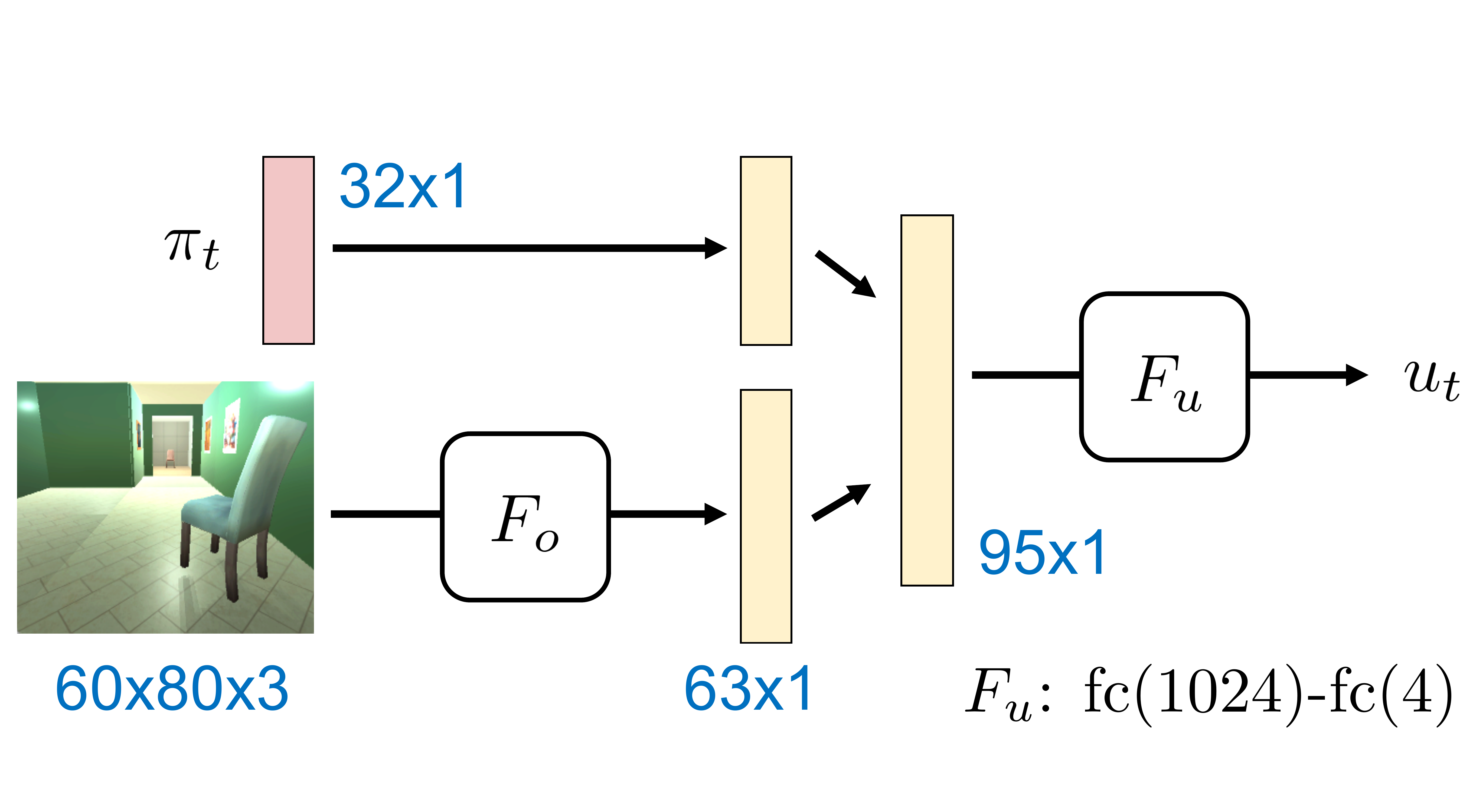}\vspace{-0.4cm}  
  \caption{The actor component maps from a camera image and a vector representation of the 
  high-level plan, $\pi_t$, to an action, $u_t$.
The $F_o$ component in \figref{fig:obsmodel} and \figref{fig:controller} has the 
same structure but the weights are not tied. 
We use $\textrm{fc}(h)$ to denote a fully connected layer with $h$ output units.
   \vspace{-6pt}} 
\label{fig:controller}
\end{figure} 

The navigation network encodes a POMDP model, 
$\dense \model(\map) = (S, A, O, T=f_{\sss T}(\cdot | \map), Z=f_{\sss
  Z}(\cdot | \map), R=f_{\sss R}(\cdot | \map)),$
which is explicitly conditioned on the floor map, $\map.$ 
 The states $s\in S$ are cells in a grid with orientation, actions $a \in A$ are
 discrete actions of the robot, and observations $o \in O$ are camera images.
 The spaces $S$, $A$ and $O$ are fixed across environments and are chosen a-priori. 
 The rewards $R=f_{\sss R}(\cdot | \map)$, 
 transition model $T=f_{\sss T}(\cdot | \map)$ and observation model
 $Z=f_{\sss Z}(\cdot | \map)$
 are conditioned on the environment, i.e. the floor map $\map.$ 
 They are represented by distinct neural network blocks, and are
 learned through end-to-end training.

 The Bayesian filter updates the belief iteratively through
\begin{equation} \label{eq:T}
\textstyle
b'_{t-1}(s) = \sum_{s'\in S}{T(s', a_{t-1}, s)b_{t-1}(s') },
\end{equation}\vspace{-0.3cm} 
\begin{equation} \label{eq:Z}
\textstyle
b_{t}(s) = \eta Z(s, o_t) b'_{t-1}(s),
\end{equation}
 where $b_t(s)$ is the belief over states $s~\in~S$ at time $t$;  $o_t \in O$ is the
 observation received after taking action $a_{t-1}~\in A$; and $\eta$ is a normalization factor.
 
 The QMDP algorithm approximates the solution to the planning problem through 
 dynamic programming. It first takes $K$ steps of value iteration,
 \vspace{-0.0cm}
\begin{equation} \label{eq:Q}
\textstyle
Q_{k+1}(s, a) =  R(s,a) + \gamma\sum_{s' \in S}{T(s, a, s') V_k(s')},
\end{equation}\vspace{-0.3cm}
\begin{equation}\label{eq:max}
\textstyle
V_k(s) = \max_a{Q_k(s,a)}.
\end{equation}
where $k = 1..K$ is the planning step, $V$ and $Q$ are state and state-action values, 
respectively.
The algorithm then computes action values, an approximation of their future value, 
\begin{equation}\label{eq:q}
\textstyle 
q(a) = \sum_{s \in S}{Q_K(s, a) b_t(s)}.
\end{equation}
 
We embed the filter and the planner in a single neural
network~(\figref{fig:navnet}) by implementing both as differentiable programs, 
\ie{} we express the algorithmic operations (\ref{eq:T} -- \ref{eq:q})
as convolutional, linear and max-pool layers.  
The operations 
are applied to the components of the POMDP model, 
$f_{\sss T}$, $f_{\sss Z}$ and $f_{\sss R}$, all of which are 
 represented by neural networks and are learned. We do not use supervision on 
 the model components. Instead, we expect a useful model to 
 emerge through training the policy end-to-end.
 
The neural network implementation of the Bayesian filter and the QMDP planner
 are shown in~\figref{fig:filterplanner}. 
 The architecture is similar to QMDP-nets~\cite{karkus2017qmdp}, 
with the notable exception of the observation model and the reactive actor component.

In our experiments the robot is restricted to a discrete grid of size $M \mytimes N$ 
and has $L$ possible orientations, where $\dense M=N=19$ and $\dense L=4$.
The input map is a $M \mytimes N$ image. 
The belief is then represented by a $M \mytimes N \mytimes L$ tensor. 
$f_{\sss T}(\cdot)$ is a single $3 \mytimes 3$ convolution with $L \cdot |A|$
output channels, one for each discrete orientation and action pair. 
$f_{\sss R}(\cdot | \map)$ is a two-layer CNN, where the input is the map $\map$ 
and the output is a $M \mytimes N \mytimes L \cdot |A|$ tensor corresponding to rewards
for each state-action pair. 

The observation model, $f_{\sss Z}$, is the most complex component of the 
learned POMDP model. 
The network architecture is show in~\figref{fig:obsmodel}. 
Unlike in QMDP-nets, we must now deal with visual observations.
 This involves inferring the environment geometry from a camera image 
and matching it against the {\twod} floor map. 
Learning the joint probability distribution $Z(s, o | \map)$ for the large 
space of image observations would be hard. Instead, we directly
represent the unnormalized conditional $z(s) = p(s | o_t, \map)$ 
by a neural network and use it directly in the belief update equation, (\ref{eq:Z}).

Navigation networks also include a low-level actor policy
 that executes the high-level plan while avoiding obstacles that the plan could not 
 account for.
 The network architecture is shown in \figref{fig:controller}.
The actor takes in a camera observation, $o_t$, and a vector representation
 of the high-level plan, $\pi_t$; and outputs a low-level action, $u_i$. In our 
 experiments $u$ and $a$ are defined in the same discrete space
 corresponding to actions in a discrete grid; but $u$
 could be continuous velocity or torque signals in future work.
The high-level plan, $\pi_t$, is represented by a vector of the computed 
$Q$ values for each high-level action, \ie{} $q_t(a)$ for $a\in A$. 
In addition, we found that a few steps of history helps the actor to avoid certain 
oscillating behaviors. Therefore, in $\pi_t$ we include four steps of past action values, 
$q_{t-\tau}(a)$, and four steps of previous action outputs,
 $u_{t-1-\tau}$, where $\tau = 1..4.$

\section{Preliminary experiments}

\begin{figure}[!t]
  \centering
	\includegraphics[width=0.95\textwidth]{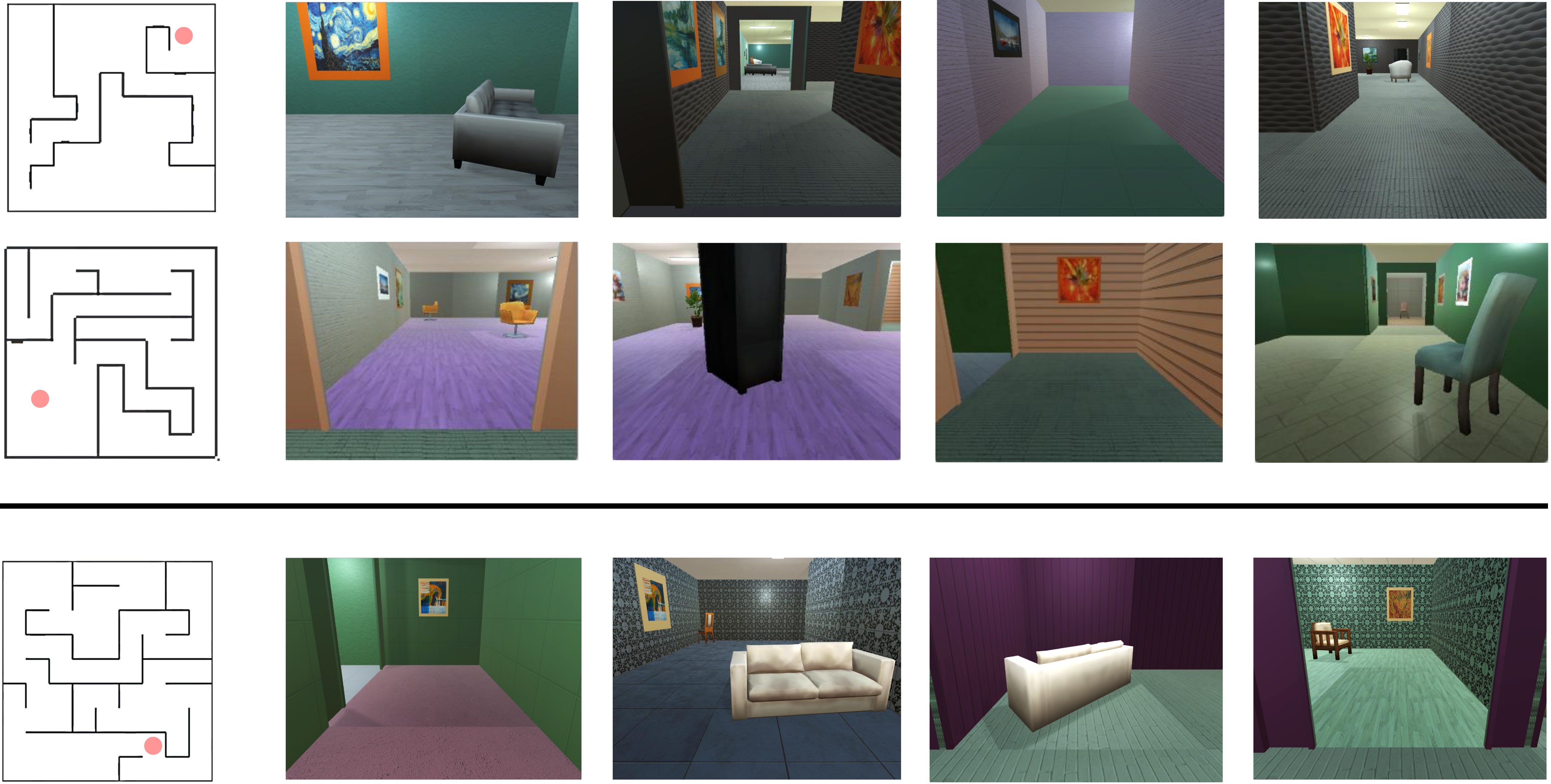}
  \caption{Examples for floor maps and camera observations from the
  training set (top) and test set (bottom).  The environment layout and 
  the textures are randomly generated. 
   \vspace{-12pt}}
\label{fig:examples}
\end{figure}

We evaluated the navigation networks for variants 
of the partially observable navigation task in a custom-built, 
high-fidelity simulator based on the Unity~3D Engine~\cite{unity3d}. 
The robot is placed in a randomly generated \threed environment, where it is
restricted to a $19\mytimes19$ discrete grid and $4$ possible orientations. 
Actions are simple motion commands: move forward, turn left, turn right,
 and stay put. The robot does not know its own state initially. Instead, it 
 receives a \twod map of the environment ($19\mytimes 19$ binary image) and 
 must estimate its location based 
 on camera observations ($60\mytimes 80$ RGB images) rendered from the
 \threed scene. 
 We place additional objects in the environment at random locations, without
  fully blocking passages.  
 The objects are picked randomly from a set of $23$ common household 
 furniture such as chairs, tables, beds, \etc{} Examples are shown in \figref{fig:examples}.
We evaluate collisions simply by assuming the robot occupies the entire grid cell it is located in.

We define three variants of the navigation task as follows:
\begin{itemize}
\item \textbf{Task A}. In this setting objects (apart from walls) 
 act solely as visual obstruction: the robot 
can go through them, and they are not shown on the 
floor map. With this tasks we evaluate the ability of localizing given camera images 
and a floor map, 
and navigating to distant goals in a new environment.
\item \textbf{Task B}. The robot can no longer move through objects, but 
the robot's map -- unlike a typical real-life floor map -- includes all objects in the
environment such as chairs, tables, etc. The robot must recognize objects and 
their relative position from camera images, and account for them for both 
localization and planning.
\item \textbf{Task C}. In this most difficult setting the robot must avoid all 
objects in the environment including furniture, while its floor map only indicates walls. 
Planning is unaware of objects not shown on the map, therefore the 
 actor may need to deviate locally from the high-level plan
 in order to avoid unanticipated obstacles.
\end{itemize}

We trained navigation networks from expert demonstrations 
in a set of $10k$ random environments. 
We then evaluated 
the learned policies in a separate set of $200$ random environments. 
For training we used successful trajectories produced 
by a clairvoyant QMDP policy, $5$ trajectories for each environment. The clairvoyant
expert has access to all objects on the map and receives binary observations
that define the occupancy of grid cells in front of the robot. 

The training was carried out by backpropagation through time using a cross-entropy 
loss between demonstrated and predicted action outputs, thus learning a policy 
end-to-end without supervision on the underlying POMDP model.
We used multiple steps of curriculum. First, we trained a policy
in synthetic grids that share the underlying structure of planning, but only
involves simple binary observations. We then trained further
 for Task~A, Task~B and Task~C in a sequence, gradually increasing the 
 difficulty\footnote{For Task~C we initialized the planner and filter weights
from Task~A, while the weights of the visual processing block in the 
actor component were initialized from Task~B.}. 
We used the full network architecture described in the previous section for 
Task~C, while for Task~A and Task~B we replaced the 
actor component by a single fully connected layer. 

\section{Results}

Preliminary results are summarized in Table~\ref{tab:all}. We report
 success rate (trials that reached the goal); 
 success rate excluding collisions (collision free trials that reached the goal);
 collision rate (trials that involved one or more collisions);
and average steps to goal (number of steps required to reach the goal, averaged for successful trials only).
Learned policies were able to successfully reach the goal in a reasonable 
number of steps with no collisions for the majority of the environments.

\floatsetup[table]{capposition=top}
\begin{table}[!t]
  \caption{Summary of results comparing navigation 
  network policies~(NavNet) and a clairvoyant QMDP policy~(QMDP).
  }
\label{tab:all}
\begin{center}
 \begin{footnotesize}
\resizebox{0.98\textwidth}{!}{%
\begin{tabular}{@{\hskip 4pt}lccc@{\hskip 4pt}}
\toprule
       & \textbf{Task A}  & \textbf{Task B}  & \textbf{Task C}  \\
{Objects on floor map} &  {walls only} & {all objects } & {walls only}  \\
{Collider objects}  & {walls only} & {all objects} & {all objects} \\
\midrule
{NavNet success rate}   &  \textbf{97.5\%}    &  \textbf{96.5\%}  &  \textbf{91.5\%} \\
	(excluding collisions)    &  (97.0\%)                 &  (95.5\%)            &  (83.0\%)  \\
{NavNet collision rate}   &  1.5\%       & 1.5\%       & 14.5\%  \\ 
{NavNet steps to goal}   		      &  37.7         & 34.4          & 36.5    \\
\midrule
{QMDP success rate}   &  81.1\%  & 84.1\%  & 62.6\%  \\
{QMDP collision rate}   &  0.0\%    & 0.0\%     & 0.0\%  \\
{QMDP steps to goal}   			  & 32.8       & 31.1      & 32.1  \\
\bottomrule 
\end{tabular}
}
\end{footnotesize}
\end{center}
\end{table}

We compared  to a clairvoyant QMDP policy that has access to much simpler 
binary observations and a map showing all obstacles; and which plans with the
``true'' underlying POMDP model. The true POMDP model would give a perfect
solution to our learning problem if the planning algorithm is exact; but not 
necessarily if the algorithm is approximate. In fact, an ``incorrect'', 
but useful model may compensate the limitations of an approximation algorithm, 
in a way similar to reward shaping in reinforcement learning~\cite{ng1999policy}.
In our experiments
learned NavNets performed significantly better than the clairvoyant QMDP. 
The QMDP algorithm is the same for both the clairvoyant QMDP and the NavNets, 
but end-to-end training allowed learning a model that is more effective for the
approximate QMDP algorithm. 
We note that the clairvoyant QMDP was used to generate the expert data for training, 
but we excluded unsuccessful demonstrations. When including both successful and unsuccessful 
demonstrations NavNets did not perform better than QMDP, as expected.

The algorithmic priors on the DNN architecture enabled policies to generalize efficiently
to previously unseen \threed environments, by carrying over the shared structure of the 
underlying reasoning required for planning.  While we defer direct comparison with
alternative DNN architectures to future work, we note that in previous reports
DNNs without algorithmic priors were unable to learn navigation policies
even in much simpler settings~\cite{karkus2017qmdp}.

While initial results are promising, the success rate and collision rate in the more
difficult setting~(Task~C) are not yet satisfactory for a real-world application. 
We observed that many failures are caused by
the inability of avoiding obstacles that were not anticipated by the plan.
 A possible reason for this is that without memory, the actor can only 
 alter the plan when an obstacle is visible. It also 
cannot choose good alternative paths locally, because it is unaware of the goal and
the plan apart from the value of the immediate next step. We may address these 
deficiencies by feeding in multiple steps of past camera observations to the 
actor, as well as a larger local ``section'' of the high-level plan.

\addtolength{\textheight}{-9cm}   

\section{Discussion}

We proposed algorithmic priors to integrate planning with deep learning.
In this section we discuss when algorithmic priors can be expected to be
effective; and identify key challenges for future research.

When comparing to standard model-free learning we should consider the following. 
Certain tasks are hard because they require a complex model to describe, however, solving 
the model can yield a simple policy. Other tasks require a complex policy, but the policy can 
be derived from a moderately complex model through tractable planning. 
Model-free policy learning can be effective in the former case, while we expect algorithmic priors to 
be critically important in the latter case.
We can apply the same reasoning when decomposing a complex problem to sub-problems: we can 
employ different degrees of algorithmic priors depending on the nature of a sub-task. 
For example, in navigation networks the filter and planner components
encode strong algorithmic priors for dealing with partial observability, but we do not 
employ algorithmic priors for the low-level actor, because this sub-task is expected to be reactive in nature.

How does the approach compare to traditional model-based learning?
Model learning is often hard because of the difficulty of inferring model
parameters from the available data; or because small model errors are
amplified through planning. Embedding the model and the algorithm in the same 
neural network allows learning end-to-end, which may in turn circumvent the
difficulties of conventional model learning.

In order to embed an algorithm in a neural network, we must implement 
it as a differentiable program. 
However, some algorithmic operations are not differentiable, 
such as indexing, sampling or argmax.
 We may deal with such operations by developing their differentiable approximations, 
 \eg{} soft-indexing in QMDP-nets~\cite{karkus2017qmdp}; or by analytically 
 approximating gradients of larger computational blocks~\cite{amos2017optnet}. 
Another concern is that repeated computation, as well as more sophisticated algorithms, 
render large neural networks, that are in turn hard to train.
Differential programs are certainly limited in terms of algorithmic computation, but
they allow learning abstract models or 
a suitable search space end-to-end, and thus may reduce the required planning computation significantly. 
Results on navigation networks and QMDP-nets demonstrate that this is indeed possible; 
however, we believe that learning more aggressive abstractions will be important 
in scaling to more difficult tasks.

In partially observable domains, a particularly important issue is the representation 
of beliefs, \ie probability distributions over states. 
Modern POMDP algorithms make planning tractable by sampling from the belief and reasoning 
with particles~\cite{silver2010monte,ye2017despot}. 
Our recent work on particle filter networks~\cite{karkus2018particle}, as well as the 
concurrent work of Jonschkowski~\etal~\cite{jonschkowski2018differentiable},
encoded particle filtering in a differentiable neural network for probabilistic state estimation. 
An exciting line of future work 
may encode more sophisticated POMDP algorithms in neural networks that plan with 
particle beliefs.

\section{Conclusion}
We proposed to integrate planning and learning through
algorithmic priors. Algorithmic priors impose the structure of planning on a DNN architecture,
by viewing DNNs as differentiable programs instead of parametric  
function approximators. Implementing all components of a larger robotic system
by differentiable programs would allow jointly optimizing the entire system for a given task.
In this paper we made a step towards this vision by encoding 
state estimation, planning and acting in a single neural network for a challenging 
partially observable navigation task. 

There are several exciting directions for future work that deal with 
high-dimensional state spaces for planning;  
encode more capable algorithms in the neural network; 
or explore how differentiable modules can be effectively pre-trained and 
transferred over domains.


\bibliographystyle{IEEEtran}
\bibliography{IEEEabrv,references}


\end{document}